\documentclass[12pt,letterpaper]{article}
\usepackage[a4paper, total={7in, 10in}]{geometry}

\usepackage[utf8]{inputenc}
\usepackage{graphicx}
\usepackage{helvet}
\usepackage{authblk}
\usepackage{hyperref}
\usepackage{amsmath} 
\usepackage{amssymb} 
\usepackage{orcidlink} 
\usepackage{csvsimple}
\usepackage[super,comma,sort&compress]  
   {natbib}
\usepackage[title]{appendix}%

\usepackage[utf8]{inputenc} 
\usepackage[T1]{fontenc}    
\usepackage{url}            
\usepackage{booktabs}       
\usepackage{amsfonts}       
\usepackage{nicefrac}       
\usepackage{booktabs}
\usepackage[table,xcdraw]{xcolor}
\usepackage{colortbl}
\usepackage{makecell}
\definecolor{LightGray}{gray}{0.95}    

\usepackage{microtype}      
\usepackage{xcolor}         
\usepackage{multirow}
\usepackage{threeparttable}
\usepackage{longtable}
\usepackage{amsmath}
\usepackage{subcaption}
\usepackage{hyperref}
\usepackage{svg}

\usepackage[table]{xcolor}
\usepackage{booktabs}
\usepackage{threeparttable}
\usepackage{helvet}   

\usepackage{xcolor}
\hypersetup{
    colorlinks=true,
    linkcolor=blue,
    filecolor=magenta,      
    urlcolor=blue, 
    pdftitle={Your Title},
}
\usepackage{makecell}
\usepackage{placeins}
\usepackage[utf8]{inputenc} 
\usepackage{booktabs}     
\usepackage{siunitx}    
\usepackage{caption}   
\usepackage{tabularx}
\usepackage{listings}    
\usepackage{pifont}
\usepackage{array}
\usepackage{makecell}
\usepackage[percent]{overpic}
\theadset{\bfseries}

\makeatletter
\renewcommand{\maketitle}{\bgroup\setlength{\parindent}{0pt}
\begin{flushleft}
  \textbf{\@title}
  
  \@author
\end{flushleft}\egroup}
\makeatother

\title{Fine-tuning an ECG Foundation Model to Predict Coronary CT Angiography Outcomes}
\date{}

\author[1,2,\#]{Yujie Xiao}
\author[3,\#]{Qinghao Zhao}
\author[1,2]{Gongzheng Tang}
\author[4]{Hao Zhang}
\author[1,2]{Zhuoran Kan}
\author[5]{Deyun Zhang}
\author[1,2]{Jun Li}
\author[1,2,6]{Guangkun Nie}
\author[1,2]{Xiaocheng Fang}
\author[1,2,7]{Haoyu Wang}
\author[1,2]{Shun Huang}
\author[4]{Tong Liu}
\author[3,*]{Jian Liu}
\author[4,*]{Kangyin Chen}
\author[1,2,8,9,*,\orcidlink{0000-0001-7521-5127}]{Shenda Hong}

\affil[1]{Institute of Medical Technology, Peking University Health Science Center, Beijing, China}
\affil[2]{National Institute of Health Data Science, Peking University, Beijing, China}
\affil[3]{Department of Cardiology, Peking University People’s Hospital, Beijing, China}
\affil[4]{Tianjin Key Laboratory of Ionic-Molecular Function of Cardiovascular Disease, Department of Cardiology, Tianjin Institute of Cardiology, The Second Hospital of Tianjin Medical University, Tianjin, China}
\affil[5]{HeartVoice Medical Technology, Hefei, China}
\affil[6]{School of Intelligence Science and Technology, Peking University, Beijing, China}
\affil[7]{University of Chinese Academy of Sciences, Beijing, China}
\affil[8]{State Key Laboratory of Vascular Homeostasis and Remodeling, NHC Key Laboratory of Cardiovascular Molecular Biology and Regulatory Peptides, Peking University, Beijing, China} 
\affil[9]{Institute for Artificial Intelligence, Peking University, Beijing, China}

\affil[$\#$]{These authors contributed equally}
\affil[*]{Correspondence: hongshenda@pku.edu.cn}

\begin{document}

\maketitle

\section*{Abstract}
Coronary artery disease (CAD) remains a major global public health burden, yet scalable pre-imaging risk stratification tools are limited. In this multicenter study, we developed and validated an artificial intelligence-enabled electrocardiography (AI-ECG) model using coronary computed tomographic angiography (CCTA) as the anatomical reference to predict vessel-specific hemodynamically significant stenosis ($\geq 70\%$ for RCA, LAD, LCX; $\geq 50\%$ for LM). The model was evaluated in internal and external cohorts, clinically normal ECGs, and prespecified demographic and clinical subgroups. It showed discrimination across vessels in internal validation and consistent external and normal ECG performance. Predicted probabilities increased with CCTA-defined stenosis severity and were converted into vessel-specific low-, intermediate-, and high-risk strata. Calibration and decision curve analyses supported its clinical utility. Integration with guideline-based pre-test probability improved risk reclassification, enhanced rule-out performance, and reduced the gray-zone proportion. In longitudinal follow-up, model-defined risk groups showed clear separation in major adverse cardiovascular events. Waveform- and attribution-based analyses identified structured ECG differences and physiologically meaningful signal regions linked to high-risk predictions. These results support AI-ECG as a feasible tool for pre-imaging risk stratification and clinical triage, warranting prospective validation in broader clinical settings.

\section*{Introduction}
Coronary atherosclerotic heart disease (CAD) is one of the leading causes of death and disability worldwide, placing a heavy burden on public health systems and socioeconomic systems \cite{cad1,cad2,cad3}. Accurately identifying the responsible vessel in CAD, assessing lesion severity, and guiding individualized treatment decisions remain core challenges in clinical practice \cite{vrints20242024}.

Currently, coronary CT angiography (CCTA) is recommended as the first-line non-invasive modality for diagnosing CAD, providing a detailed assessment of coronary anatomy and stenosis severity, while serving as a crucial tool for pre-procedural planning and strategy guidance for percutaneous coronary intervention (PCI)\cite{ccta,ccta2}. Nevertheless, the widespread application of CCTA is limited by its dependence on advanced imaging infrastructure and post-processing techniques, strict requirements for heart rate control and respiratory coordination, radiation exposure, and the risk of contrast-induced nephropathy, particularly in patients with impaired renal function \cite{narula2020scct,jiliang,kidney}. These limitations underscore the need for complementary, low-cost, and non-invasive approaches that can assist in CAD risk assessment and triage.

Electrocardiography (ECG), as a non-invasive, convenient, and real-time detection method, has unique advantages in the early identification and risk stratification of cardiovascular diseases. Changes in ECG waveforms, such as ST segment shift and T wave inversion, have been widely used in the preliminary diagnosis of acute coronary syndrome (ACS)\cite{ECGCAD1,ECGCAD2,ECGCAD3}. Meanwhile, with the rapid development of artificial intelligence (AI), especially deep learning technology, AI is being widely used in clinical research \cite{ai4s1,ai4s2,ai4s3,ai4s4,ai4s5,ai4s6}. Utilizing AI to perform deep feature mining of ECG signals to predict the pathogenesis of cardiovascular diseases has become a research hotspot \cite{research1,research2,research3,research4,research5}. However, existing AI-ECG studies using CCTA as the reference standard have mainly focused on patient-level screening for obstructive coronary artery disease, with limited attention to vessel-specific prediction of hemodynamically significant coronary stenosis. For example, Choi et al.'s study mainly targeted the detection of obstructive coronary artery disease\cite{choi2023electrocardiogram}, while Park et al.'s study focused on predicting obstructive lesions in patients with stable angina\cite{park2024artificial}. Although several studies have used ECG and AI models to localize myocardial infarction (MI), these studies mainly targeted infarct location or scar-related substrates after prior MI, rather than CCTA-defined vessel-specific stenosis in patients undergoing routine coronary evaluation, particularly among individuals without overt ECG abnormalities\cite{wang2025artificial}.

To overcome these limitations, we conducted a multicenter study using clinical data from Peking University People's Hospital and the Second Hospital of Tianjin Medical University, together with an independent longitudinal follow-up cohort. We adopted a transfer learning framework to fine-tune ECGFounder, a large-scale pretrained ECG foundation model \cite{ecgfounder1,ecgfounder2}, which had been further adapted by our group on an independent dataset for MI-related prediction tasks. Based on this framework, we developed an AI-ECG model to directly predict hemodynamically significant vessel-specific coronary artery stenosis defined by CCTA from standard ECG signals. The model identified hemodynamically significant stenosis in the RCA, LAD, and LCX, defined as luminal narrowing of at least 70\%, and clinically significant LM stenosis, defined as luminal narrowing of at least 50\%. Importantly, the model provided vessel-specific predictions rather than only patient-level assessment.

In this study, model performance was systematically evaluated using receiver operating characteristic analysis, subgroup analysis, and comparisons of predicted probability distributions across different CCTA-defined stenosis severities. To assess whether the model retained clinical relevance in patients without obvious abnormalities on routine ECG interpretation, we further evaluated its performance in the subgroup of clinically normal ECGs. In addition, we translated continuous model-predicted probabilities into vessel-specific risk strata according to predefined sensitivity- and specificity-based thresholds. The clinical utility of this risk stratification strategy was examined using calibration analysis, decision curve analysis, and in the follow-up cohort, longitudinal Kaplan-Meier analysis.

Beyond model evaluation, we further investigated whether AI-derived risk strata could complement established clinical risk assessment. Specifically, we designed a fusion strategy that integrated model-defined risk categories with guideline-based PTP categories and compared its clinical performance with that of the AI model alone and PTP assessment alone. Finally, to support clinical interpretation, we conducted waveform-based and attribution-based analyses to characterize ECG morphology differences across model-defined risk groups and to identify the signal regions contributing most strongly to model prediction.

We position the proposed AI-ECG model along a tiered framework of clinical applications. The primary intended use, directly supported by the present study, is pre-imaging risk stratification in symptomatic patients with suspected CAD who are being considered for non-invasive anatomical evaluation. The broader aspirational use is to extend AI-ECG-based triage to earlier decision nodes in the evaluation of suspected CAD, including primary care, outpatient cardiology, and emergency department settings, where access to CCTA or expert cardiology consultation may be limited. While the current retrospective evidence is derived from a CCTA-referred population, the model's preserved discrimination in patients with apparently normal ECGs and across heterogeneous clinical subgroups suggests potential for broader application, which we explicitly frame as requiring prospective validation in lower-prevalence cohorts.

\section*{Results}
\subsection*{Baseline Character of Dataset}
\begin{figure}[htbp]
    \centering
    \captionsetup{font=small, labelfont=bf}

    \includegraphics[width=1.01\textwidth]{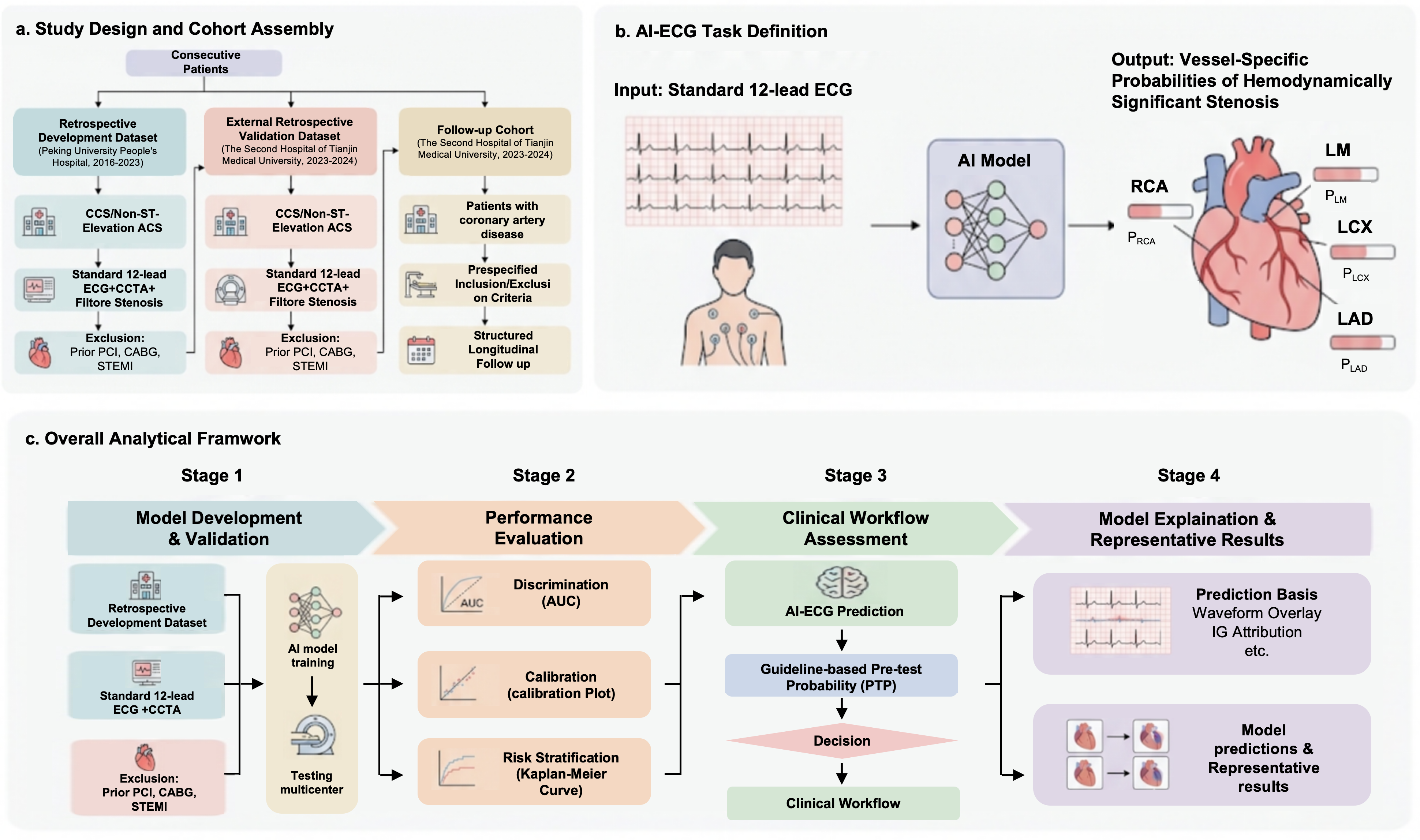}

    \caption{
    \textbf{Overall study design, AI-ECG task definition, and analytical framework.}
    \textbf{a)} Study Design and Cohort Assembly. Flowchart detailing the consecutive patient inclusion and exclusion criteria across three distinct cohorts: a retrospective development dataset from Peking University People's Hospital, an external retrospective validation dataset from the Second Hospital of Tianjin Medical University, and a dedicated longitudinal follow-up cohort.
    \textbf{b)} AI-ECG Task Definition. Illustration of the model's primary objective, where standard 12-lead ECG waveforms are utilized as the sole input to predict vessel-specific probabilities of hemodynamically significant stenosis across four major coronary arteries: the left main (LM), left anterior descending (LAD), left circumflex (LCX), and right coronary artery (RCA).
    \textbf{c)} Overall Analytical Framework. The study pipeline is divided into four stages: (Stage 1) AI model development and multicenter validation; (Stage 2) comprehensive performance evaluation assessing discrimination (AUC), calibration, and risk stratification (Kaplan-Meier analysis); (Stage 3) clinical workflow assessment integrating AI predictions with guideline-based pre-test probability (PTP) for actionable decision-making; and (Stage 4) model interpretation using waveform overlay analysis and Integrated Gradients-based attribution.
    }
    \label{fig:framework}
\end{figure}

The study was built on consecutive multicenter clinical data from two centers, comprising 4,620 internal ECG-CCTA pairs from Peking University People's Hospital, 2,477 external validation cases from the Second Hospital of Tianjin Medical University, and an additional 400-patient longitudinal cohort from the Second Hospital of Tianjin Medical University. This design supports model development, independent external evaluation, and follow-up-based clinical assessment while reducing the risk of arbitrary case selection. The workflow in Figure \ref{fig:framework} further separates retrospective model development, external validation, and longitudinal clinical assessment, providing a coherent structure for evaluating discrimination, calibration, risk stratification, clinical workflow integration, and model interpretation.

Baseline age distributions were broadly comparable across datasets. The mean age was 64.5 $\pm$ 10.4 years in the internal dataset, 63.1 $\pm$ 11.7 years in the external validation dataset, and 63.14 $\pm$ 10.43 years in the follow-up cohort. The proportion of patients aged $\geq65$ years was also similar between the external validation dataset and the follow-up cohort, at 50.5\% and 50.2\%, respectively, whereas the corresponding proportion in the internal dataset was 43.0\%. The internal dataset contained 746 missing records for both sex and age, corresponding to 16.1\% of the cohort, whereas missingness was absent or negligible in the external validation dataset and the follow-up cohort.

Cardiometabolic comorbidity patterns varied across cohorts without showing a uniform direction of risk enrichment. Diabetes was more common in the external validation dataset than in the internal dataset, at 35.9\% versus 31.0\%, but was lower in the follow-up cohort at 24.3\%. Hypertension showed a different pattern, with the highest prevalence in the follow-up cohort at 57.2\%, compared with 52.2\% in the internal dataset and 50.7\% in the external validation dataset. Prior myocardial infarction remained uncommon across all datasets, ranging from 2.0\% in the external validation dataset to 4.2\% in the internal dataset and 4.0\% in the follow-up cohort. These differences indicate that external validation was performed under moderate shifts in baseline clinical risk rather than under an identical case mix.

Across all datasets, LAD had the highest prevalence of hemodynamically significant stenosis among the four vessel-specific tasks, reaching 29.7\% in the internal dataset, 31.0\% in the external validation dataset, and 20.8\% in the follow-up cohort. RCA and LCX showed lower but clinically meaningful disease burdens, with RCA-positive rates of 17.2\%, 25.6\%, and 16.0\%, and LCX-positive rates of 17.2\%, 23.0\%, and 13.8\% in the internal, external, and follow-up datasets, respectively. This distribution indicates that vessel-specific modeling was not driven by a single uniformly prevalent endpoint, but instead required learning across tasks with different positive class frequencies.

LM stenosis showed the lowest prevalence in the external validation and follow-up cohorts, with hemodynamically significant stenosis present in only 89 of 2,477 external cases (3.6\%) and 6 of 400 follow-up patients (1.5\%). Although the internal dataset contained a higher LM-positive proportion of 10.3\%, the marked reduction in the external and longitudinal cohorts indicates substantial class imbalance for this task. Consequently, LM-specific performance estimates, calibration, and downstream risk stratification should be interpreted cautiously.

Detailed baseline information is provided in Supplement Table A1. And the time difference between ECG and CCTA is shown in Supplement Figure A1.

\begin{table}[htbp]
\centering
\begin{threeparttable}
\caption{Baseline Characteristics of Internal and External Datasets and Cohort}
\label{tab:baseline_character}

\definecolor{zebra}{HTML}{F2F2F2}
\sffamily \footnotesize 
\renewcommand{\arraystretch}{1.3} 

\newcolumntype{Y}{>{\centering\arraybackslash}X}

\rowcolors{5}{white}{zebra}

\setlength{\tabcolsep}{4pt} 

\begin{tabularx}{\textwidth}{l Y Y Y}
\arrayrulecolor{black}\toprule[1.5pt] 

\rowcolor{zebra}
\textbf{Characteristic} & 
\textbf{Peking University People's Hospital (Internal)} & 
\textbf{The Second Hospital of Tianjin Medical University (External)} & 
\textbf{Cohort of the Second Hospital of Tianjin Medical University}\\

\rowcolor{zebra} 
 & ($n=4,620$) & ($n=2,477$) & ($n=400$)\\ 

\arrayrulecolor{black!30}\midrule[0.5pt] 

\textbf{Demographics} & & \\
\quad Sex & & & \\
\quad \quad Male & 1,952 (42.3\%) & 1,283 (51.8\%) & 183 (45.8\%) \\
\quad \quad Female & 1,922 (41.6\%) & 1,194 (48.2\%) & 217 (54.2\%) \\
\quad \quad Missing & 746 (16.1\%) & 0 (0.0\%) & 0 (0.0\%) \\

\quad Age (years) & & & \\
\quad \quad Mean $\pm$ SD & 64.5 $\pm$ 10.4 & 63.1 $\pm$ 11.7 & 63.14 $\pm$ 10.43 \\
\quad \quad $<65$ & 1,886 (40.8\%) & 1,223 (49.4\%) & 199 (49.8\%) \\
\quad \quad $\geq65$ & 1,988 (43.0\%) & 1,252 (50.5\%) & 201 (50.2\%) \\
\quad \quad Missing & 746 (16.1\%) & 2 (0.1\%) & 0 (0.0\%) \\

\textbf{Medical History} & & & \\
\quad Diabetes & & & \\
\quad \quad Yes & 1,433 (31.0\%) & 889 (35.9\%) & 97 (24.3\%) \\
\quad \quad No & 3,187 (69.0\%) & 1,588 (64.1\%) & 303 (75.7\%) \\

\quad Hypertension & & & \\
\quad \quad Yes & 2,411 (52.2\%) & 1,256 (50.7\%) & 229 (57.2\%) \\
\quad \quad No & 2,209 (47.8\%) & 1,221 (49.3\%) & 171 (42.8\%) \\

\quad Prior myocardial infarction & & & \\
\quad \quad Yes & 193 (4.2\%) & 49 (2.0\%) & 16 (4.0\%) \\
\quad \quad No & 4,427 (95.8\%) & 2,428 (98.0\%) & 384 (96.0\%) \\

\textbf{Coronary Artery Stenosis} & & & \\
\quad RCA & & & \\
\quad \quad Hemodynamically Significant Stenosis & 795 (17.2\%) & 635 (25.6\%) & 64 (16.0\%) \\
\quad \quad Hemodynamically Non-significant Stenosis & 3,825 (82.8\%) & 1,842 (74.4\%) & 336 (84.0\%) \\

\quad LM & & & \\
\quad \quad Hemodynamically Significant Stenosis & 477 (10.3\%) & 89 (3.6\%) & 6 (1.5\%) \\
\quad \quad Hemodynamically Non-significant Stenosis & 4,143 (89.7\%) & 2,388 (96.4\%) & 394 (98.5\%) \\

\quad LAD & & & \\
\quad \quad Hemodynamically Significant Stenosis & 1,373 (29.7\%) & 769 (31.0\%) & 83 (20.8\%) \\
\quad \quad Hemodynamically Non-significant Stenosis & 3,247 (70.3\%) & 1,708 (69.0\%) & 317 (79.2\%) \\

\quad LCX & & & \\
\quad \quad Hemodynamically Significant Stenosis & 795 (17.2\%) & 570 (23.0\%) & 55 (13.8\%) \\
\quad \quad Hemodynamically Non-significant Stenosis & 3,825 (82.8\%) & 1,907 (77.0\%) & 345 (86.2\%) \\

\arrayrulecolor{black!10}\bottomrule 
\end{tabularx}

\begin{tablenotes}
    \scriptsize \sffamily 
    \item Data are presented as n (\%) or mean $\pm$ SD. 
    \item \textit{Abbreviations}: SD, standard deviation; RCA, right coronary artery; LM, left main coronary artery; LAD, left anterior descending artery; LCX, left circumflex artery.
    \item  Hemodynamically significant stenosis was defined as $\geq 70\%$ stenosis for RCA, LAD, and LCX, and as $\geq 50\%$ stenosis for LM.
\end{tablenotes}
\end{threeparttable}
\end{table}

\subsection*{Development and Multi-Center Validation of AI-ECG Model}
\begin{figure}[htbp]
    \centering
    \captionsetup{font=small, labelfont=bf}

    \includegraphics[width=1.01\textwidth]{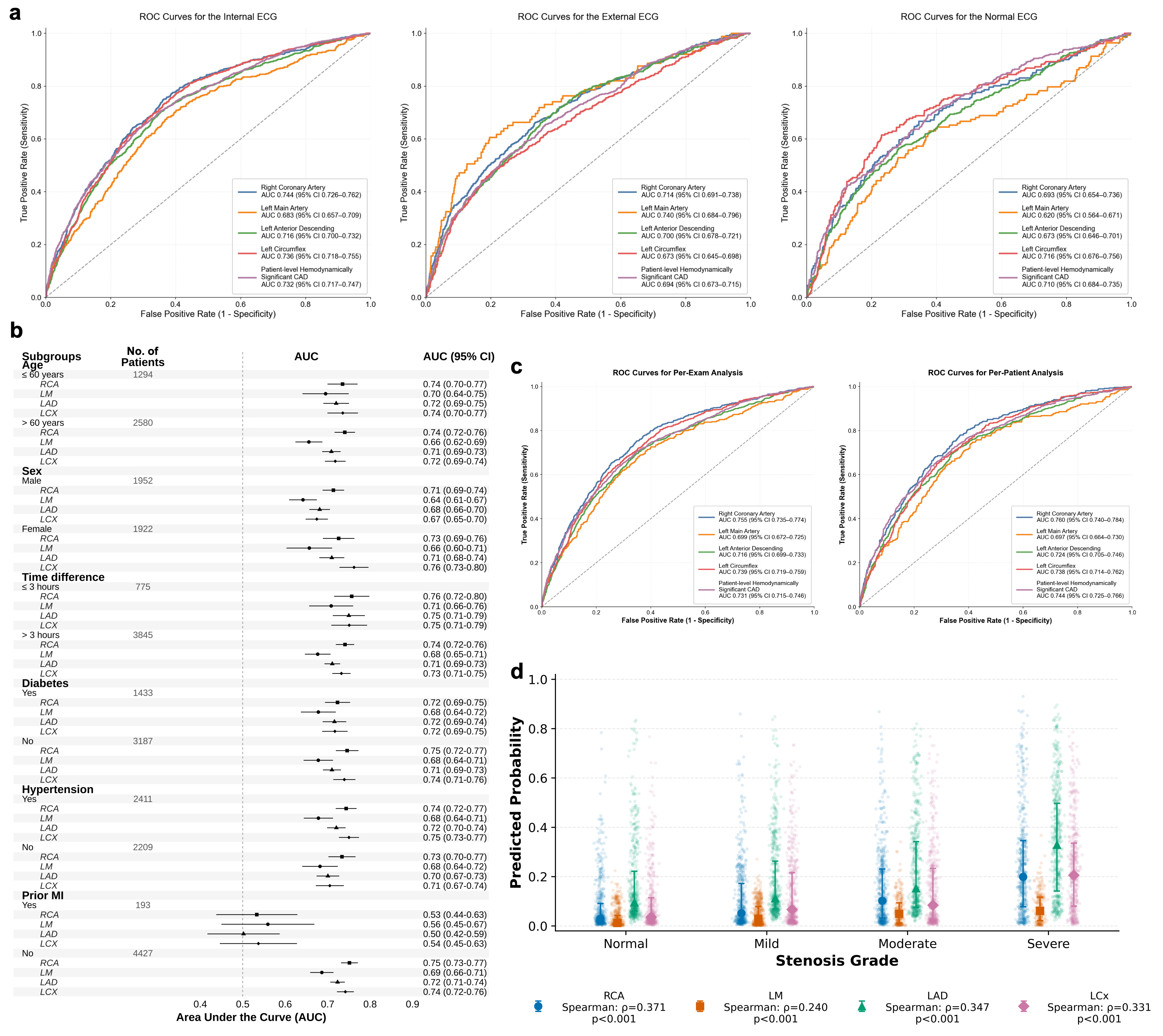}
    \caption{
    \textbf{Validation of AI-ECG for CCTA-Defined Coronary Stenosis.}
    \textbf{a)} ROC curves for patient-level hemodynamically significant CAD and vessel-specific prediction of hemodynamically significant stenosis in the internal validation cohort, external validation cohort, and individuals with clinically normal ECGs.
    \textbf{b)} Subgroup analysis of model discrimination across age, sex, ECG-CCTA time interval, diabetes, hypertension, and prior myocardial infarction, with AUC and 95\% CI reported for each coronary artery.
    \textbf{c)}  ROC curves at the CCTA examination level and patient level, evaluating the consistency of model performance across different analytical granularities.
    \textbf{d)}  Distribution of model-predicted probabilities across CCTA-defined stenosis grades, showing the agreement between continuous model outputs and disease severity for RCA, LM, LAD, and LCX.
    }
    \label{fig:validation}
\end{figure}

We adopted a five-fold stratified cross-validation strategy, with patient identifiers used as the grouping variable, to ensure that all ECG records from the same individual were assigned to a single fold and that the distribution of multi-label coronary artery outcomes remained balanced across folds. To comprehensively evaluate the generalizability of the model across multicenter datasets and its performance in individuals with clinically normal ECGs, we reported the micro-averaged area under the receiver operating characteristic curve (micro-AUC) for the internal dataset and evaluated the external validation dataset using an ensemble of models derived from the five cross-validation folds. Model performance was further assessed in the subgroup of patients with clinically normal ECG interpretations (Figure \ref{fig:validation} a).

The results showed that the AI-ECG model achieved an area under the curve (AUC) of 0.732 for predicting patient-level hemodynamically significant CAD in the internal validation set. Patient-level hemodynamically significant CAD was defined as luminal stenosis $\geq 70\%$ in any major coronary branch or stenosis $\geq 50\%$ in the left main coronary artery. For individual vessels, the AUC values for the RCA, LM, LAD, and LCX were 0.744, 0.683, 0.716, and 0.736, respectively. In the external validation set, the model achieved an AUC of 0.694 for detecting patient-level hemodynamically significant CAD, with corresponding vessel-specific AUC values of 0.714, 0.740, 0.700, and 0.673. These findings indicate that the AI-ECG model provided stable discriminative performance for identifying both patient-level hemodynamically significant CAD and vessel-specific hemodynamically significant stenosis. The external validation results were broadly consistent with the internal validation results, further supporting the robustness of model generalization across multicenter datasets.

We further evaluated predictive performance in individuals with clinically normal ECGs, defined as ECGs interpreted as normal in physician diagnostic reports. In this challenging subgroup, the model maintained an AUC of 0.710 for identifying patient-level hemodynamically significant CAD, with vessel-specific AUC values of 0.693 for RCA, 0.620 for LM, 0.673 for LAD, and 0.716 for LCX. These results were broadly consistent with those observed in the full population, indicating that the AI-ECG model can detect the risk of patient-level hemodynamically significant CAD and vessel-specific hemodynamically significant stenosis even in individuals with apparently normal ECGs, thereby highlighting its potential value for identifying CCTA-defined coronary stenosis risk among patients with suspected CAD despite apparently normal routine ECG interpretations.

In addition, we calculated the 95\% confidence interval for the AUC of each vessel. These intervals further support the statistical reliability and stability of model discrimination.

To account for potential differences between examination-level and patient-level evaluation, we additionally performed ROC analyses at both the CCTA examination level and the patient level, with the corresponding curves shown in Figure \ref{fig:validation} c.

We further performed subgroup analyses to evaluate model stability across population characteristics and acquisition timing. After stratification by age ($<65$ years versus $\geq65$ years), sex, coronary artery stenosis status, the interval between ECG and CCTA acquisition ($\leq 3$ hours versus $>3$ hours), and medical history, the model showed broadly comparable discriminative performance across subgroups, without evidence of substantial performance degradation.

As shown in Figure \ref{fig:validation} b, model performance remained largely consistent across prespecified subgroups, with no substantial decline in AUC, except among individuals with a prior history of myocardial infarction.

Specifically, higher AUC values were observed in the subgroup with an ECG-CCTA acquisition interval of 3 hours or less, as well as among younger individuals and those with a history of hypertension. Model discrimination was comparable between male and female participants and did not differ materially according to diabetes status.

Collectively, these findings suggest that the proposed model demonstrates stable discriminative performance across diverse population characteristics and testing conditions, supporting its potential applicability in heterogeneous clinical settings.

Supplementary Table A2 provides subgroup analysis results stratified by characteristics such as degree of coronary artery stenosis, medication use, presence of calcified plaques, arrhythmia type, and chief complaint.

To further assess the association between model-predicted probabilities and the actual severity of coronary stenosis, we generated a jittered strip plot with median and percentile error bars, as shown in Figure \ref{fig:validation} d. This plot shows the distribution of predicted probabilities for the four major coronary arteries across four stenosis grades: normal, mild, moderate, and severe. All vessels showed a consistent monotonic increase, indicating that the risk probabilities assigned by the model increased with greater stenosis severity.

Spearman rank correlation analysis further confirmed a statistically significant positive association between predicted probabilities and stenosis severity for all coronary arteries ($p < 0.001$). The strongest correlation was observed for the right coronary artery (RCA; $\rho = 0.371$). Notably, the absolute predicted probabilities for the left main coronary artery (LM; range, 0.0-0.4) were substantially lower than those for the other vessels, which may reflect the low prevalence of LM lesions in the training population. Nevertheless, the model retained significant discriminative ability for LM ($\rho = 0.240$).

\subsection*{Model's Decision-making Ability in Coronary Artery Stenosis Risk Stratification}
This study focused on four major coronary arteries and defined two risk thresholds for each vessel according to the sensitivity and specificity of the corresponding prediction model. Patients were subsequently stratified into high-risk, intermediate-risk, and low-risk groups. Specifically, the threshold corresponding to a sensitivity of 90 percent was used to separate the low-risk and intermediate-risk groups, whereas the threshold corresponding to a specificity of 95 percent was used to separate the intermediate-risk and high-risk groups.

\begin{figure}[htbp]
    \centering
    \captionsetup{font=small, labelfont=bf}

    \includegraphics[width=1.01\textwidth]{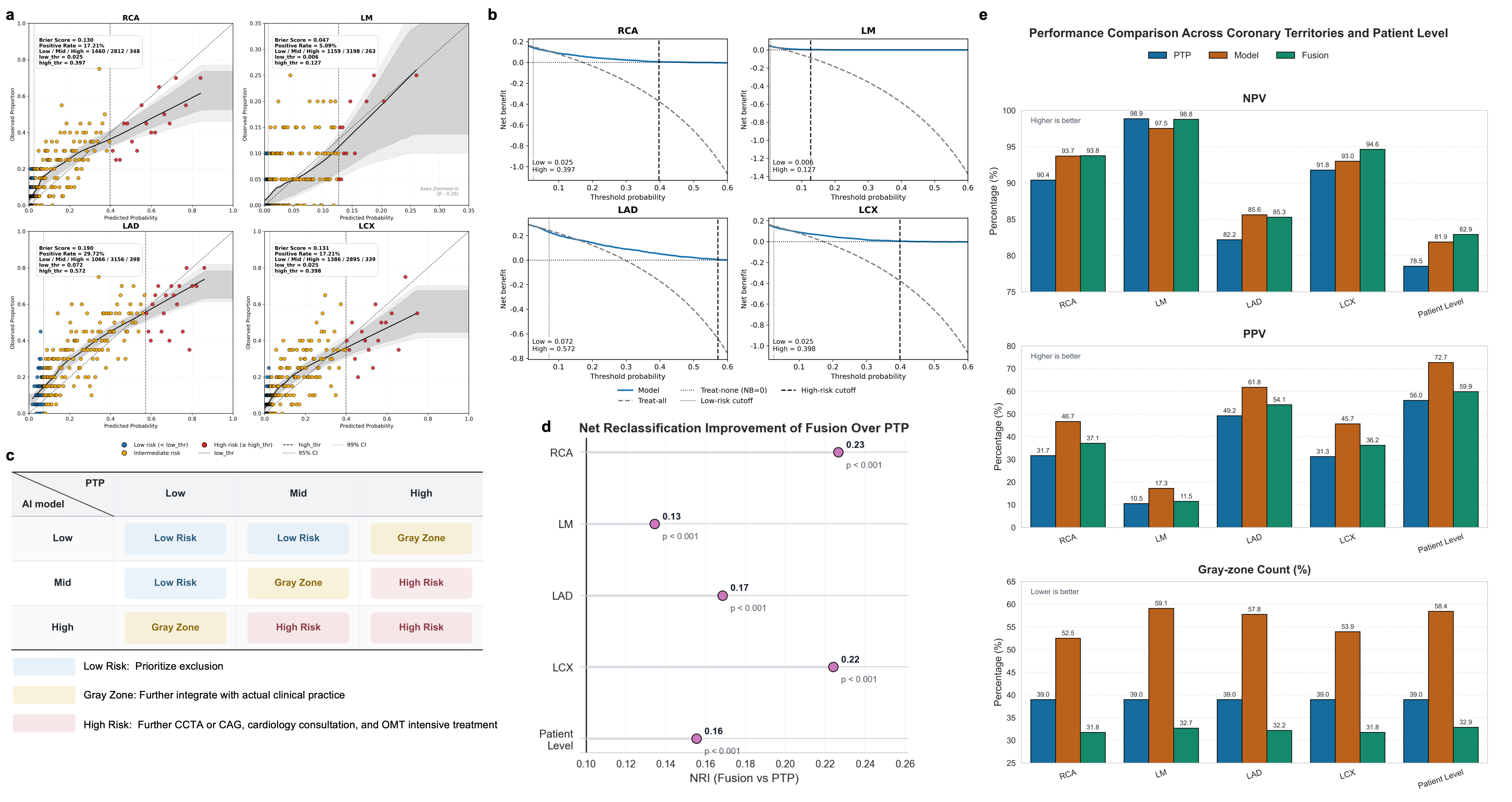}
    \caption{
    \textbf{Clinical Utility of AI-ECG-Guided Risk Stratification for Coronary Stenosis.}
    \textbf{a)} Calibration plots for RCA, LM, LAD, and LCX, showing the agreement between predicted and observed risk and supporting the accuracy of the model-derived low-, intermediate-, and high-risk grouping.
    \textbf{b)} Decision curve analysis for RCA, LM, LAD, and LCX, showing the net clinical benefit of model-based risk grouping across threshold probabilities and supporting its clinical consistency.
    \textbf{c)} Final clinical decision scheme integrating AI model risk groups with guideline-based PTP categories to define low-risk, gray-zone, and high-risk groups for downstream management.
    \textbf{d)} Net reclassification improvement of the final fusion strategy over traditional PTP at the patient level and across coronary territories, with corresponding p values.
    \textbf{e)} Comparison of NPV, PPV, and gray-zone proportion among PTP, the AI model, and the fusion strategy across coronary territories and at the patient level.
    }
    \label{fig:utility}
\end{figure}

To assess the reliability of the model predictions, we plotted calibration curves for the prediction results, as shown in Figure \ref{fig:utility} a.

The calibration curves indicate that the model provided reliable predictions for all four coronary arteries. The Brier scores for RCA, LM, LAD, and LCX were 0.130, 0.047, 0.190, and 0.131, respectively. Notably, for hemodynamically significant stenosis of the LM, the model did not produce very high predicted probabilities because of the extremely low prevalence of this lesion in the general population, which is consistent with its epidemiological distribution.

To evaluate the potential clinical utility of the proposed risk stratification strategy, we further conducted decision curve analysis (DCA) to quantify the net benefit of model-guided decision-making across clinically relevant risk thresholds. As shown in Figure \ref{fig:utility} b, for the four major coronary arteries, the DCA curves of the model generally lay above the treat-all and treat-none strategies, indicating that model-based risk stratification can provide greater net clinical benefit within the relevant threshold ranges and supporting its potential value as an auxiliary decision-making tool.

Furthermore, we integrated model-based risk stratification with PTP-based stratification according to the ESC guidelines \cite{19guideline}, and designed the clinical risk grouping strategy shown in Figure \ref{fig:utility} c. According to the guidelines, PTP scores were divided into low-risk, intermediate-risk, and high-risk groups using thresholds of 5\% and 15\%. These groups were then combined with the low-risk, intermediate-risk, and high-risk groups defined by the model. Specifically, the final low-risk group included patients classified as low risk by the model and as low or intermediate risk by PTP, as well as patients classified as intermediate risk by the model and as low risk by PTP. The final high-risk group included patients classified as intermediate risk by the model and as high risk by PTP, as well as patients classified as high risk by the model and as intermediate or high risk by PTP. All remaining patients were assigned to the final gray zone group.

Based on this strategy, we plotted the bar charts shown in Figure \ref{fig:utility} e, which present the NPV at the low-risk threshold, the PPV at the high-risk threshold, and the observed incidence rate within the gray zone group at both the patient level and the four-vessel level. At the four-vessel level, these results were derived by combining model predictions with PTP scores. At the patient level, a patient was classified as high risk if any of the four vessels was classified as high risk by the model, and as low risk only if all four vessels were classified as low risk. This patient-level model classification was then combined with the PTP score to obtain the final result.

As shown in Figure \ref{fig:utility} e, the proposed fusion strategy improved NPV relative to both the model alone and PTP alone. Although PPV decreased relative to the model alone, it remained higher than that of PTP alone. This pattern is likely attributable to the inherently low PPV of PTP, which reduced the overall PPV of the fusion strategy. Furthermore, our fusion strategy feasibly reduces the number of people in the gray zone compared to using AI models or PTP alone. As shown in Figure \ref{fig:utility} d, the fusion strategy yielded positive Net Reclassification Improvement (NRI) values relative to PTP, and all results were statistically significant ($p < 0.001$). These findings indicate that the proposed model can preserve the improvement in NPV while accepting a moderate reduction in PPV in exchange for a lower gray zone count. Therefore, the model shows strong potential as a screening and triage tool. Nevertheless, definitive diagnosis still requires further integration into routine clinical practice. In addition, predictions for high-risk vessels, particularly LM and LAD, warrant closer clinical attention. Beyond binary screening for overall coronary disease, the model's capacity to localize risk to specific vascular territories—particularly the LM and LAD—carries substantial clinical utility. In real-world triage pathways, an AI-ECG alert indicating a high probability of severe LM or LAD disease can directly inform pre-procedural optimization. It empowers clinicians to prioritize urgent functional or anatomical imaging, escalate pre-procedural medical management, and alert interpreting radiologists to closely scrutinize heavily calcified or ambiguous segments in the flagged territories during subsequent CCTA evaluation.

\subsection*{Future Risk Prediction}
\begin{figure}[htbp]
    \centering
    \captionsetup{font=small, labelfont=bf}

    \includegraphics[width=0.8\textwidth]{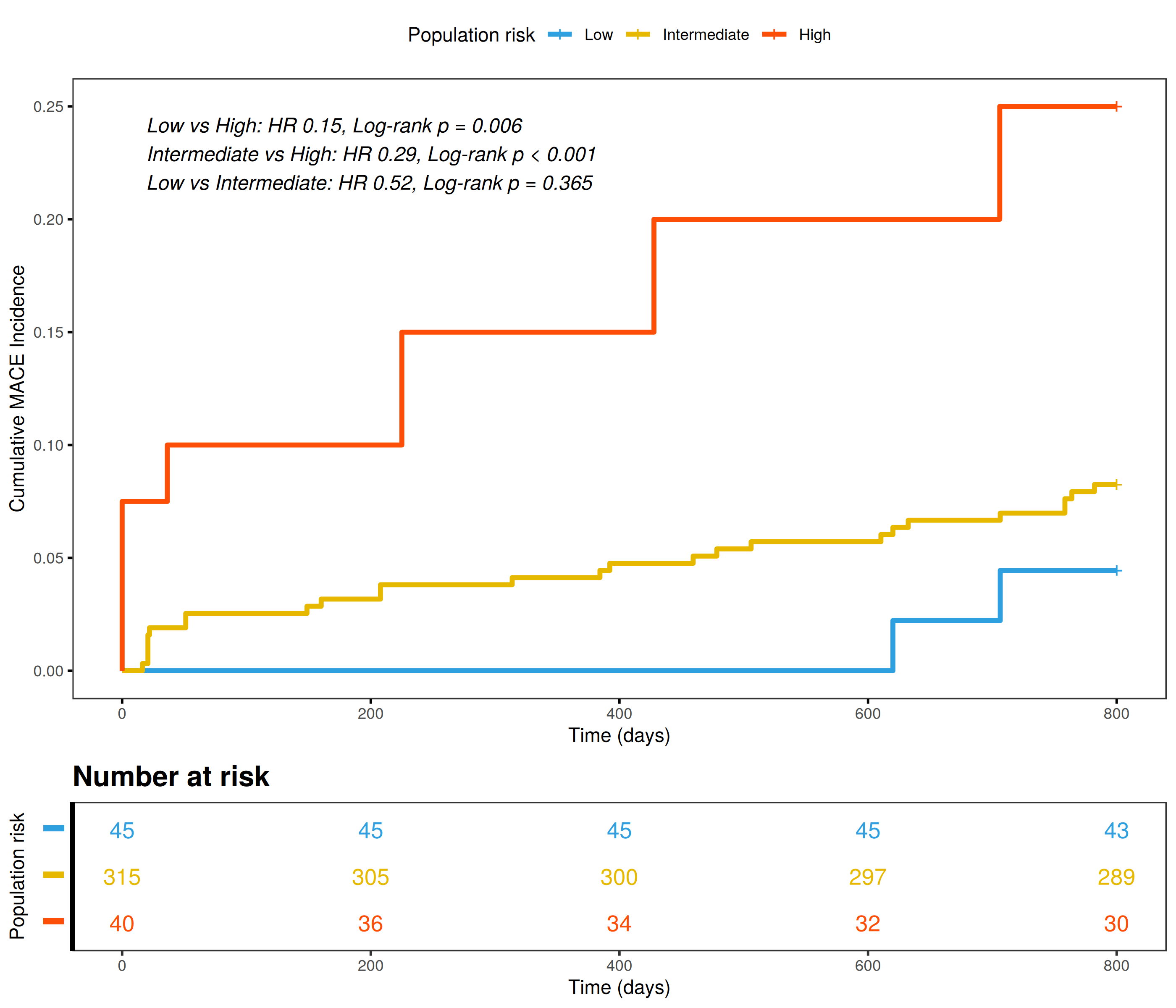}

    \caption{\textbf{Patient-Level Kaplan–Meier Curves Stratified by Model-Predicted Risk.}
    Kaplan–Meier curves showing the cumulative incidence of major adverse cardiovascular events (MACE) over the follow-up period among patients stratified into low-, intermediate-, and high-risk groups according to patient-level AI-ECG risk classification. The number-at-risk table is shown below the plot.
    }
    \label{fig:km_curve}

    \vspace{8pt}
\end{figure}

To further evaluate the value of the model for predicting future clinical outcomes and stratifying risk, we assessed the event incidence at preset follow-up time points based on patient-level risk groupings predicted by the model, as shown in Figure \ref{fig:km_curve}. The ECG acquisition time was used as the starting point for follow-up, and an event was defined as the occurrence of a major adverse cardiovascular event (MACE) during follow-up. MACE included myocardial infarction (MI), death, or revascularization by PCI or CABG.

The analysis focused on events occurring within the predefined 800-day follow-up window. Using Kaplan-Meier analysis, we compared the longitudinal MACE risk among the high-risk, intermediate-risk, and low-risk groups during follow-up, thereby evaluating the ability of the model to distinguish longitudinal changes in the risk of future MACE.

As shown in Figure \ref{fig:km_curve}, the Kaplan-Meier curves demonstrated a clear gradient in cumulative MACE incidence across the three population-level risk groups over the 800-day follow-up period. The component events included 31 incident MI, 8 deaths, and 42 revascularization procedures. The follow-up duration was sufficient to support the prespecified 800-day observation window, with a lower quartile of 947 days, an upper quartile of 992 days, and a mean duration of 1050.7 days. The high-risk group consistently exhibited the highest event rate, with an early increase in cumulative incidence, indicating that individuals classified as high risk by the model were more likely to experience adverse cardiovascular events even in the early stage of follow-up. The intermediate-risk group showed an event rate between those of the high-risk and low-risk groups, with a gradual increase over time. In contrast, the low-risk group maintained a very low cumulative event incidence during most of the follow-up period, with only a slight increase observed at later time points, suggesting that the proposed stratification strategy has good ability to identify individuals at low near-term risk.

Pairwise comparisons further supported the discriminative value of this risk stratification. Compared with the high-risk group, both the low-risk group and the intermediate-risk group showed significantly lower MACE risk. Specifically, the comparison between the low-risk group and the high-risk group yielded a hazard ratio (HR) of 0.15 with a log-rank $p = 0.006$, whereas the comparison between the intermediate-risk group and the high-risk group yielded an HR of 0.29 with a log-rank $p < 0.001$.

The overall results suggest that the proposed population-level risk stratification can feasibly distinguish individuals at substantially elevated future MACE risk, thereby supporting its potential value for longitudinal risk assessment and clinical follow-up management.

\subsection*{Explanation of the Model's Results}
\begin{figure}[htbp]
    \centering
    \captionsetup{font=small, labelfont=bf}

    \includegraphics[width=1.0\textwidth]{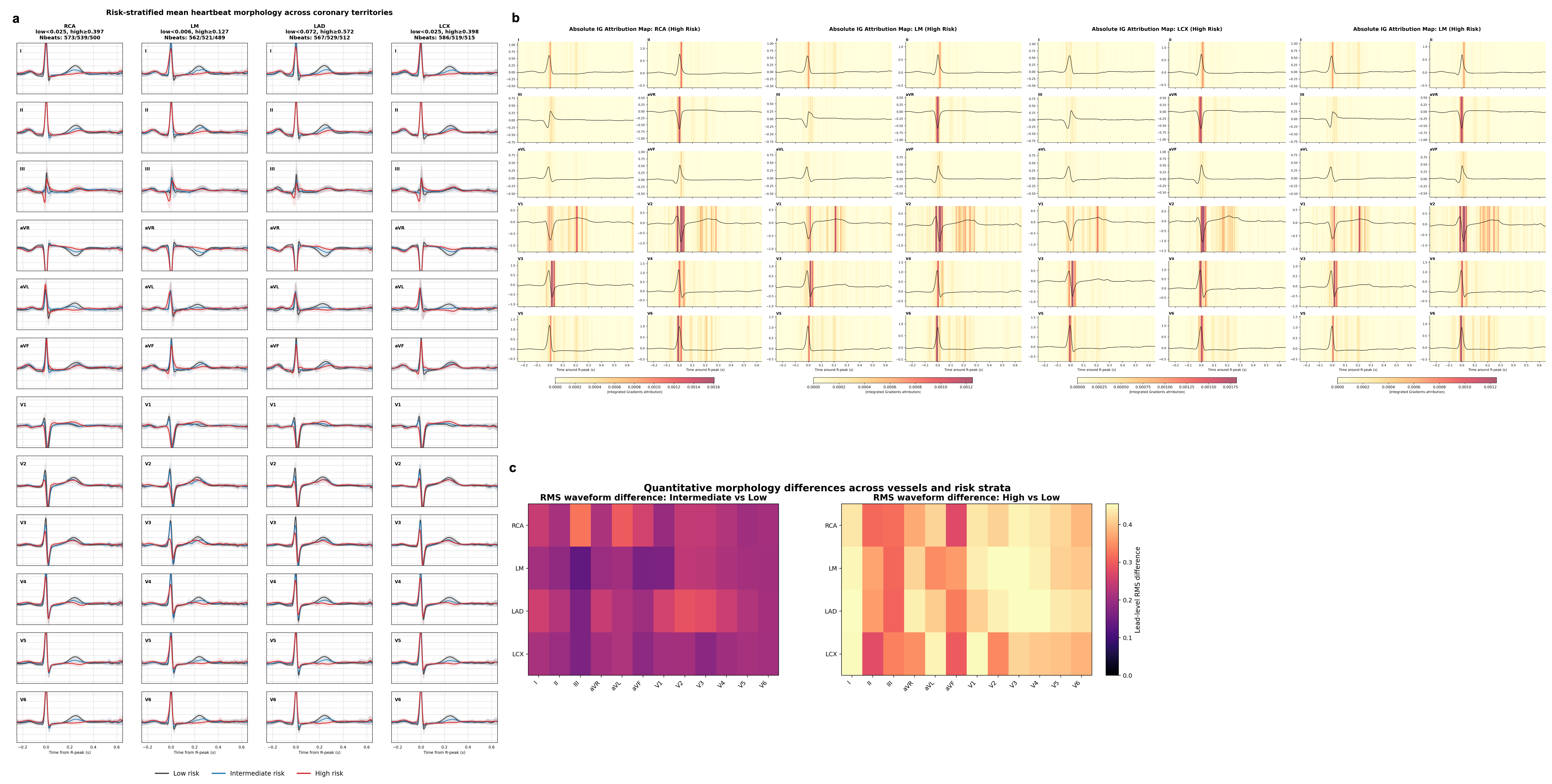}

    \caption{\textbf{Explanation of the model's results.}
    \textbf{a)} Mean heartbeat morphology for low-, intermediate-, and high-risk groups based on model probability grouping. \textbf{b)} Lead diagnostic attribution plots highlighting ECG regions most associated with high-risk prediction. \textbf{c)} Heatmaps of RMS waveform differences across leads and coronary territories, comparing intermediate-risk versus low-risk and high-risk versus low-risk groups. In panel a, colors denote risk groups; in panel c, color intensity represents the magnitude of waveform difference.}
    \label{fig:visualize}
    \vspace{8pt}
\end{figure}

To further elucidate the electrophysiological basis of model-driven risk stratification, we aligned heartbeat cycles to the R peak and compared the mean morphology across low-, intermediate-, and high-risk strata for each coronary territory. As shown in Figure \ref{fig:visualize} a, panel a presents group-level averaged heartbeat morphologies after stratifying the entire cohort according to model-predicted risk. The separation among the three risk groups is not driven by random fluctuations in isolated leads, but by reproducible morphology shifts that remain visible across multiple leads within each vessel-specific task. Overall, the high-risk group exhibits the largest deviation from the low-risk template, whereas the intermediate-risk group typically lies between the two, indicating that the predicted risk ordering is reflected in a graded change of waveform morphology rather than an abrupt binary transition.

The attribution maps in Figure \ref{fig:visualize} b further show that the model does not distribute importance uniformly over the entire beat. Panel b was obtained by selecting the 50 samples with the highest predicted risk for each vessel and averaging their attribution responses. The resulting maps show that salient contributions are temporally concentrated around structured ECG intervals, with the densest attribution bands appearing near the QRS complex and its adjacent repolarization-related regions. This localization pattern is observed across the four vessel-specific models, although the exact temporal emphasis differs by coronary territory. Such behavior is important for interpretability, because it indicates that the classifier is anchored to physiologically meaningful waveform segments rather than to diffuse background noise or visually incidental fluctuations. In this sense, the discrimination mechanism of the model appears to be driven by localized morphologic differences embedded within clinically recognizable portions of the cardiac cycle.

The quantitative heatmaps in Figure \ref{fig:visualize} c support the waveform-level observations by summarizing the average morphology difference at the population level across the full database. Specifically, panel c presents cohort-level RMS difference heatmaps, showing that the morphology difference between the high-risk and low-risk groups is consistently larger than that between the intermediate-risk and low-risk groups across most vessels and leads. The intermediate-versus-low comparison remains detectable but is visibly weaker and more heterogeneous, which is consistent with the expectation that intermediate-risk individuals occupy a transitional morphologic regime. By contrast, the high-versus-low comparison yields broader and stronger differences, especially across multiple precordial leads, indicating that the extreme ends of the model-defined risk spectrum correspond to more pronounced ECG remodeling. Importantly, this quantitative pattern matches the visual impression from Figure \ref{fig:visualize} a and the localized attribution evidence from Figure \ref{fig:visualize} b, forming a coherent explanation chain in which larger inter-group morphology shifts are associated with clearer model emphasis on physiologically structured waveform segments.

\section*{Discussion}
This study constructed and validated an artificial intelligence model based on ECG for the prediction of hemodynamically significant coronary stenosis as shown by CCTA. Building upon the large-scale pre-trained ECG foundation model ECGFounder, the framework was further adapted to tasks related to coronary artery disease through transfer learning strategies. Consequently, this adaptation extends existing AI-ECG research beyond patient-level coronary risk assessment by enabling vessel-specific prediction of hemodynamically significant stenosis, which may provide more localized and clinically actionable information for pre-imaging triage.

The cohort design supports the clinical relevance of the proposed pre-imaging triage task. All datasets included patients undergoing routine CCTA, including CCS, UA, or NSTEMI, while excluding those with STEMI. This exclusion criteria ensured that the study population closely resembles the demographic for whom routine CCTA is commonly considered in real-world practice. Furthermore, the requirement of ECG acquisition prior to CCTA and exclusion of prior PCI or CABG reduced the risk that electrical changes following intervention or clinical decisions informed by imaging could influence the input of the model. Therefore, the task evaluated in this study represents not a retrospective reconstruction of established anatomical disease post-treatment, but rather a clinically meaningful scenario wherein ECG is utilized before anatomical confirmation to support the prioritization of examinations.

For model development and anatomical label construction, CCTA was used as the reference examination. Although invasive CAG remains the reference standard for diagnosing coronary artery disease, CCTA was selected as the primary anatomical reference in this study for several reasons. In real-world clinical practice, patients referred for CAG usually have more evident clinical manifestations or a higher pre-test probability of obstructive coronary artery disease, which may introduce spectrum bias and limit the evaluation of an ECG-based tool intended for pre-imaging risk stratification and triage in patients with suspected CAD. In contrast, CCTA is widely used as a non-invasive anatomical test in patients with suspected coronary artery disease and has a high negative predictive value for excluding hemodynamically significant stenosis. Therefore, for a decision-support tool designed to comprehensively identify coronary stenosis patterns from ECG and to assist clinical risk stratification before anatomical imaging, CCTA provides a more suitable reference standard for cohort construction and model evaluation.

An analysis of the baseline characteristics indicates that the datasets preserved real-world cardiometabolic complexity rather than representing an artificially selected population with low risk. The internal and external datasets exhibited similar age distributions, and the follow-up cohort maintained a comparable mean age and a balanced age structure. Diabetes and hypertension were prevalent across all cohorts, although the prevalence of these conditions was lower in the follow-up cohort compared to the development and external validation datasets. This pattern introduces a modest distributional shift while simultaneously retaining clinically expected comorbidity profiles. The stability of downstream analyses under this shift suggests that the proposed framework may remain applicable beyond a single, narrowly defined development population.

The inclusion of patients with prior myocardial infarction further augmented clinical realism, although this subgroup concurrently highlighted a challenging source of heterogeneity. A plausible explanation is that some recorded histories of prior myocardial infarction may reflect myocardial infarction with non-obstructive coronary arteries (MINOCA) or prior infarctions that did not undergo revascularization, thereby leaving no documented history of PCI or CABG. Moreover, while prior myocardial infarction was uncommon and thus unlikely to dominate the learning process of the model, persistent ECG abnormalities following an infarction may attenuate the direct correspondence between current electrical patterns and stenosis defined by CCTA. This phenomenon elucidates the reduced performance within this subgroup and emphasizes that prior myocardial injury may constitute a distinct electrophysiological background rather than merely presenting simple noise. Consequently, the composition of the baseline supports broad clinical applicability while concurrently identifying patient groups that necessitate dedicated evaluation.

Regarding predictive performance, the model demonstrated clinically significant discrimination for the vessel-specific prediction of coronary stenosis. Within the internal dataset, the framework achieved stable AUCs across the RCA, LM, LAD, and LCX, and the vessel-specific performance was comparable to that of the overall obstructive CAD model. This comparability suggests that the network did not merely learn a non-specific signal indicative of global coronary disease, but instead captured information capable of being mapped to distinct coronary territories. The preservation of this discriminative ability in the external dataset further substantiates the generalizability of the signal. Collectively, these findings indicate that AI based on ECG can advance beyond binary screening for coronary disease toward risk assessment that is more anatomically informative.

The performance observed in clinically normal ECGs suggests that the model successfully captured subtle electrophysiological information that remains largely inaccessible to routine visual interpretation. Because patients with visually normal ECGs are expected to manifest fewer overt ischemic abnormalities, this subgroup constitutes a stringent test for latent features derived from ECG. In this setting, the model retained meaningful discrimination. Therefore, the proposed approach may provide supplementary value for patients whose standard ECG interpretation does not reveal clear ischemic changes. Furthermore, the predicted probabilities increased monotonically alongside the severity of stenosis defined by CCTA across all vessels. This graded relationship bolsters the interpretation that the output of the model reflects the burden of disease rather than random classification artifacts.

The outcomes of risk stratification demonstrate that continuous probabilities from the model can be translated into clinically interpretable decision categories. By establishing the low-risk cutoff under a constraint of 90\% sensitivity and the high-risk cutoff under a constraint of 95\% specificity, the model was aligned with two complementary clinical objectives: safe rule-out and confident rule-in. Calibration analyses revealed plausible agreement between the predicted risk and the observed frequency of hemodynamically significant stenosis. Simultaneously, these analyses indicated that tasks with low prevalence, particularly those involving the LM, require cautious interpretation. This evidence supports the utilization of model probabilities not solely as ranking scores, but also as continuous estimates of risk that can inform clinical decision thresholds.

Decision curve analysis further corroborates the clinical utility of risk stratification assisted by the model. Across threshold ranges of clinical relevance, the framework generally provided a higher net benefit compared to treat-all and treat-none strategies, particularly for the RCA, LAD, and LCX. This pattern suggests that the implementation of the model can curtail unnecessary downstream testing or intervention while preserving the capacity to identify patients who may derive benefit from further evaluation. For the LM, the diminished net benefit remains consistent with the low event prevalence and the narrower range of prediction associated with this vessel. Consequently, the decision curve analysis implies that the model holds practical value for risk-guided triage, while also delineating the boundary conditions under which interpretations should exercise greater conservatism.

The evaluation of the fusion strategy demonstrates that risk strata derived from AI can complement, rather than replace, clinical assessment based on guidelines. Crucially, the core advantage of this AI-ECG framework lies in its zero-radiation, cost-effective, and highly accessible nature. While advanced risk reclassification often incorporates anatomical metrics like the Coronary Artery Calcium Score (CACS), obtaining a CACS inherently necessitates existing CT scanning infrastructure. Our model was explicitly designed for primary care and emergency settings where such imaging resources are unavailable or where patients are unsuitable for radiation exposure. In these resource-constrained scenarios, the AI-ECG framework serves as a low-barrier frontline triage tool—providing incremental risk stratification beyond PTP while being significantly more accessible than imaging-dependent prerequisites (CACS). By integrating model-defined risk groups with the widely established ESC PTP categories, the proposed decision matrix prioritized concordant low-risk cases for rule-out, clinically concerning high-risk cases for escalation, and uncertain cases for further evaluation. This structural design mirrors the practical application of risk tools, wherein model outputs must be synthesized with pre-existing clinical probabilities rather than being interpreted in isolation. The subsequent improvement in NPV and the corresponding reduction in the proportion of the gray zone indicate that this fusion approach can render risk stratification more actionable and highly compatible with existing workflows.

Reclassification analyses yield additional evidence that the fusion strategy contributes information beyond the utilization of PTP alone. Positive NRI values were observed at both the patient level and the vessel level, yielding statistically significant improvements across all comparisons. These findings suggest that the fusion framework did not merely reproduce guideline-based risk categories, but actively reallocated patients toward classifications of events and non-events that are more appropriate. Overall, the fusion approach establishes a more balanced operating point compared to the reliance on AI prediction or PTP assessment independently, effectively transforming the absence of baseline imaging data into a strategic demonstration of the model's low-barrier, high-impact clinical utility.

Subsequent follow-up analysis implies that the proposed system for risk stratification captures prognostic information that extends beyond cross-sectional anatomical stenosis. In the independent follow-up cohort, patients classified as high risk exhibited the highest cumulative incidence of MACE, whereas patients categorized as low and intermediate risk demonstrated substantially lower hazards. This distinct separation indicates that vessel-level risk derived from ECG may reflect not only the immediate presence of obstructive disease but also a broader paradigm of cardiovascular vulnerability. Therefore, the developed model may support both examination prioritization at the baseline and longitudinal risk management subsequent to the initial diagnostic encounter.

Notably, the decision curve analysis, Kaplan-Meier analysis, and calibration assessment in this study were primarily conducted to evaluate the clinical relevance and decision-support potential of the AI model itself. The further clinical value of the proposed AI-assisted decision framework, particularly when integrated with guideline-based PTP assessment, should be confirmed in prospective studies before routine clinical implementation.

The primary innovations of this study encompass several key advancements. A foundational contribution is the capability of the proposed model to directly extract clinically relevant information from raw ECG signals to predict clinically significant vessel-specific coronary stenosis as defined by CCTA. Furthermore, the model facilitates vessel-specific risk assessment for the RCA, LM, LAD, and LCX, rather than merely predicting the presence of coronary lesions. Notably, the model maintained robust performance even on ECGs interpreted as normal in routine clinical practice, suggesting the capture of latent electrophysiological features that remain difficult to identify through visual interpretation. This finding underscores the potential value of AI-ECG analysis for the screening of occult coronary artery disease. Building upon this predictive capability, the study systematically evaluated whether continuous prediction probabilities can be converted into clinically actionable risk strata. A dual-threshold strategy, based on sensitivity and specificity, enabled the definition of low-, intermediate-, and high-risk groups tailored to distinct clinical objectives. Calibration curves, decision curve analysis, and the graded increase in predicted probabilities across CCTA-defined stenosis severities demonstrated that the outputs of the model provide both statistical discrimination and clinically interpretable risk information. Consequently, the model was evaluated not solely as a classifier, but also as a tool for decision support. A further significant contribution is the integration of AI-derived risk strata with the PTP score recommended by the ESC guidelines, proposing a fusion-based decision scheme aligned with real-world clinical workflows. Within this scheme, low-risk patients are prioritized for rule-out, high-risk patients are referred for intensive evaluation or intervention, and intermediate-risk patients remain in a gray zone that necessitates further judgment based on clinical information. Compared with the use of the PTP score alone or the AI model alone, the fusion strategy demonstrated incremental value by improving low-risk rule-out, reducing the proportion of the gray zone, and achieving a positive NRI. These results indicate that the fusion strategy enhances the clinical interpretability of model outputs and strengthens the compatibility with existing guideline-based pathways. Ultimately, MACE analysis in the follow-up cohort indicated that this risk stratification system reflects not only the current anatomical stenosis burden defined by CCTA, but also the risk of future adverse cardiovascular events. Together, these findings provide a comprehensive evidence chain for the use of ECG for pre-imaging risk stratification, examination prioritization, and longitudinal risk assessment in patients with suspected coronary artery disease.

The intended use of the proposed AI-ECG model spans a layered clinical pathway. The directly validated use case is pre-imaging triage among symptomatic patients with suspected CAD who are being considered for CCTA, corresponding to the population on which the model was developed. In this setting, the fusion strategy with guideline-based PTP demonstrated improved rule-out performance, reduced gray-zone proportion, and positive net reclassification improvement, supporting its incremental value over standard clinical assessment.

Beyond this immediate use case, the long-term clinical aspiration is to deploy AI-ECG-based triage at earlier decision points in the evaluation of suspected CAD — including primary care, general outpatient cardiology, and emergency departments — where ECG is widely available but advanced imaging access is limited. Two observations from the present study support, but do not yet prove, the feasibility of this extension. First, the model retained meaningful discriminative ability in patients whose 12-lead ECGs were interpreted as normal in routine clinical practice, suggesting that it captures sub-clinical electrophysiological information not readily accessible to visual interpretation. Second, the model maintained broadly stable performance across heterogeneous demographic and clinical subgroups, indicating that its decision signal is not narrowly tied to a single phenotype.

However, the present retrospective evidence is derived from a CCTA-referred population with a relatively high pre-test probability of obstructive coronary disease. In lower-prevalence populations such as primary care, the absolute negative predictive value would be expected to increase further, while the positive predictive value would correspondingly decrease. Calibration may also shift across patient mixes with different cardiometabolic profiles. Therefore, the broader application of this AI-ECG framework to upstream triage settings requires dedicated prospective evaluation in cohorts representative of the intended deployment environment, including primary care patients with chest pain symptoms, emergency department patients with suspected non-ST-elevation acute coronary syndromes, and outpatient cardiology populations with suspected stable angina.

Although this study demonstrates methodological innovation and explores the clinical applicability of AI-enabled ECG analysis for coronary artery disease, several limitations should be acknowledged. First, although the model was developed and externally validated using data from two medical centers, larger and more diverse cohorts are needed to further evaluate its generalizability across different populations, institutions, and ECG acquisition settings. In addition, both the development and external validation cohorts consisted of patients referred for CCTA, representing a clinically enriched population with suspected coronary artery disease. Therefore, prevalence-dependent metrics such as PPV and NPV, as well as the predefined risk thresholds, may not be directly transferable to lower-prevalence populations, including lower-prevalence outpatient or primary-care populations with suspected CAD. Prospective recalibration and validation will be necessary before applying the model beyond CCTA-referred populations. Second, the relatively low prevalence of hemodynamically significant stenosis in the LM may have introduced class imbalance and affected the stability of performance estimates for this vessel-specific task. Therefore, the LM results should be interpreted with caution.

Several limitations are also related to cohort composition and clinical applicability. Patients with prior PCI or CABG were excluded from the training and validation datasets, which limits the applicability of the proposed model for evaluating in-stent restenosis or graft vessel occlusion after revascularization. In addition, the model showed reduced performance among individuals with a history of myocardial infarction, possibly because persistent post-infarction ECG alterations obscure lesion-specific electrophysiological patterns. Dedicated studies involving post-revascularization populations and patients with prior myocardial infarction are therefore warranted. Moreover, because the cohort was derived from patients undergoing routine CCTA, the study population may not fully represent patients referred directly for emergency CAG. This exclusion limits the ability of the model to distinguish ECG patterns associated with acute versus chronic myocardial infarction.

Methodologically, CCTA was used as the anatomical reference standard. Although CCTA has high clinical value for non-invasive assessment of coronary artery disease, diagnostic discrepancies may exist between CCTA-defined stenosis and invasive CAG findings, particularly in heavily calcified lesions. Although we conducted a subgroup analysis according to the presence of coronary calcified plaques, only binary descriptions of plaque presence or absence were available, and quantitative calcification scores were not recorded. This precluded further analyses based on coronary artery calcium score burden.

Another limitation concerns the clinical decision fusion strategy, which was constructed using the 2019 ESC pre-test probability categories. We acknowledge that the 2024 ESC Guidelines recommend a Risk Factor-weighted Clinical Likelihood model and emphasize the role of the CACS in risk reclassification. In our retrospective, real-world multicenter datasets, systematic calcium scores and comprehensive longitudinal risk factor profiles were not uniformly available for all patients, precluding direct implementation of the 2024 criteria. However, rather than diminishing the model's value, this data constraint authentically reflects the very clinical reality the model targets: early triage for patients who have not yet undergone any anatomical imaging. The 2019 PTP model remains a well-established and clinically validated baseline for demonstrating the proof of concept of our AI-assisted fusion strategy in such pre-imaging scenarios. Furthermore, the proposed framework is modular and may be adapted to newer risk stratification pathways when more granular multimodal clinical data, including calcium scores, become available in future prospective cohorts.

Finally, the relationship between model predictions and downstream clinical decision-making requires further investigation. Although model-guided risk stratification showed potential clinical utility, its impact on diagnostic pathways, patient outcomes, and resource allocation should be evaluated in prospective studies. The explainability analyses also require cautious interpretation. While these analyses highlighted ECG regions that contributed to model predictions, the clinical significance of subtle signal-level variations that are not readily discernible by human interpretation remains uncertain. Moreover, current explainability methods have inherent methodological limitations, underscoring the need for further refinement and validation.

Future research will incorporate large-scale data from additional centers to further enhance model stability and generalizability. In parallel, more refined analyses of LM coronary lesions will be conducted. This includes the integration of spatial information from ECG to better characterize electrocardiographic patterns associated with distinct anatomical regions and to improve lesion-specific identification. Importantly, to clarify the relationship between model performance and clinical decision-making, a prospective randomized controlled trial is currently being designed. This trial will evaluate the impact of AI-ECG-guided risk stratification on downstream diagnostic strategies and clinical outcomes in real-world settings.

\section*{Methods}
\subsection*{Overall Study Design and Methodological Framework}
This study aimed to construct an AI-ECG model to predict coronary artery lesions shown by CCTA based on ECG, thus providing a potential auxiliary tool for non-invasive risk assessment of CAD. The model was developed based on 2595 CCTA examinations and 4620 paired ECG data from 2323 patients at Peking University People's Hospital. To systematically evaluate the model's generalization ability across different centers and populations, we further validated it externally on 2477 CCTA examinations and 2477 paired ECG data from 2477 patients at the Second Hospital of Tianjin Medical University. In addition, this study included a longitudinal follow-up cohort of 400 patients from the Second Hospital of Tianjin Medical University to assess the model's prospective risk stratification ability in real-world clinical scenarios.

To comprehensively evaluate the clinical applicability of the model, this study establishes a systematic validation framework covering discrimination, risk stratification, clinical decision support, prognostic relevance, and interpretability (As shown in Figure \ref{fig:framework}). First, receiver operating characteristic curves are used to assess the ability of the model to identify CCTA-defined hemodynamically significant vessel-specific stenosis. This evaluation is conducted in the internal validation dataset, the external validation dataset, the dataset with ECGs interpreted as normal in routine clinical practice, and at both the examination and patient levels, to examine the robustness of the model across data sources and analytic units. We then compare the distribution of predicted probabilities across different CCTA-defined stenosis grades to assess whether the model captures the graded severity of coronary artery disease. Subgroup analyses are further performed across demographic characteristics, medical history, the time interval between ECG and CCTA examinations, arrhythmia type, medication use, and clinical symptoms to evaluate the consistency and generalizability of model performance across clinical populations.

On this basis, the study defines low-, intermediate-, and high-risk groups using probability thresholds derived from predefined sensitivity and specificity targets, and evaluates the agreement between these risk strata and the observed frequency of CCTA-defined lesions using calibration curves. To further assess the potential clinical value of the model, decision curve analysis is performed across different risk thresholds to compare the net benefit of the model-assisted strategy with treat-all and treat-none strategies. To explore how the model can be integrated into clinical practice, this study combines AI-derived risk strata with the pre-test probability score recommended by the ESC guidelines to construct a fusion-based decision scheme. The AI-only strategy, PTP-only strategy, and fusion strategy are then compared in terms of negative predictive value, positive predictive value, gray-zone proportion, and net reclassification improvement over the guideline-based strategy.

To evaluate the ability of the model to identify future adverse events at an early stage, vessel-level risk predictions are further aggregated into patient-level risk categories in the follow-up cohort. Kaplan-Meier survival analysis and the log-rank test are then used to assess differences in future major adverse cardiovascular events across risk groups. Together, this evaluation pipeline extends from cross-sectional identification of anatomical stenosis to clinical decision support and longitudinal prognostic stratification, providing a complete and coherent evidence chain for validating the clinical performance of the model.

In addition, to examine the interpretability of model predictions, this study compares the morphology of averaged single-beat waveforms across different risk groups in the 12-lead ECG and uses attribution analysis to identify the key leads and temporal segments that contribute to prediction. This analysis helps provide an intuitive understanding of the potential mechanisms by which AI-enabled ECG detects coronary artery disease, and offers a basis for subsequent clinical interpretation and methodological refinement.

\subsection*{Data Sources}
This multicenter study included clinical data from Peking University People's Hospital and the Second Hospital of Tianjin Medical University. To minimize potential selection bias and ensure the representativeness of the study population, we adopted a consecutive inclusion strategy rather than arbitrary sampling. Specifically, we systematically queried the complete electronic medical record (EMR) systems, electrocardiogram information management systems, picture archiving and communication systems (PACS), and cardiac catheterization imaging systems of the participating hospitals to identify all potentially eligible patients during the predefined study periods.

The internal development dataset was constructed from all consecutively eligible patients who underwent CCTA at Peking University People's Hospital between 2016 and 2023. After applying the predefined eligibility criteria, this dataset included 2,323 patients, 2,595 CCTA examinations, and 4,620 paired electrocardiogram (ECG) recordings. In the internal development dataset, a single CCTA examination was allowed to be matched with multiple eligible ECG recordings acquired before the CCTA examination. This strategy was adopted to increase the diversity of ECG presentations, expand the effective training sample size, and improve the ability of the model to recognize severe coronary lesions from ECGs without overt diagnostic abnormalities. For external validation, we applied the same consecutive inclusion and exclusion criteria to the complete clinical database of the Second Hospital of Tianjin Medical University between 2023 and 2024. To ensure a strict temporal correspondence between ECG and CCTA findings in the external validation setting, each CCTA examination was matched only with the nearest ECG recording obtained on the same day before the CCTA examination. This external validation dataset comprised 2,477 patients, 2,477 CCTA examinations, and 2,477 paired ECG records.

The exclusion criteria for the CCTA-based model development and validation cohorts were designed to reduce potential confounding from previous or subsequent coronary revascularization procedures and acute treatment pathways. Patients with prior PCI or CABG related to the indexed CCTA examination were excluded. Patients with ST-segment elevation myocardial infarction (STEMI) were also excluded because current guidelines recommend immediate reperfusion therapy, including primary PCI, for patients presenting with STEMI \cite{ECGCAD1}. Consequently, STEMI patients are less likely to undergo CCTA before revascularization in routine clinical practice, resulting in limited and highly selected CCTA data for this population. Moreover, STEMI represents an acute coronary occlusion phenotype with distinct ECG manifestations and treatment pathways. Accordingly, the primary CCTA-based study population consisted of patients with chronic coronary syndrome (CCS), unstable angina (UA), and non-ST-segment elevation myocardial infarction (NSTEMI).

To further minimize the potential influence of coronary revascularization procedures or post-CCTA clinical interventions on ECG characteristics, all paired ECG recordings were required to have been acquired before the corresponding CCTA examination. When multiple ECG recordings were available for a given CCTA examination, ECGs obtained before CCTA were matched according to the predefined temporal matching strategy. This combined temporal and clinical exclusion strategy was adopted to reduce bias introduced by procedure-related or treatment-related ECG alterations.

In addition to the CCTA-based development and validation cohort, we constructed a longitudinal follow-up cohort to assess the model's ability to stratify the risk of future adverse clinical events. This cohort consecutively included patients who visited the Department of Cardiology at the Second Hospital of Tianjin Medical University with CAD between January 1, 2023, and December 31, 2024. Inclusion criteria were: (1) age $\geq18$ years; (2) underwent a standard 12-lead electrocardiogram (ECG). Exclusion criteria were: (1) severe ECG noise, artifacts, or technical errors rendering the ECG unanalyzable; (2) in-hospital death during the current hospitalization. After applying these criteria, a total of 400 consecutive patients were included in the longitudinal follow-up cohort.

Informed consent was obtained from all eligible patients in the longitudinal follow-up cohort. Clinical data, including medical history, medication use, laboratory tests, and other examination results, were extracted from the structured electronic medical record system. ECG data were obtained from an independent ECG information management system. Telephone follow-ups were conducted every six months. The primary endpoint was MACE, defined as a composite of MI, all-cause death, PCI, or CABG.

The study was approved by the Peking University Medical Ethics Committee (Approval No.: IRB00001052-23071), the Ethics Committee of Peking University People's Hospital (Approval No.: 2024PHB428-001), and the Ethics Committee of the Second Hospital of Tianjin Medical University (Approval No.: KY2025K386). For the retrospective CCTA-based development and validation cohorts, the requirement for informed consent was waived in accordance with relevant ethical guidelines because of the retrospective nature of the study and the use of de-identified clinical data. For the longitudinal follow-up cohort, written informed consent was obtained from all eligible patients before inclusion.

\subsection*{Labeling Strategy}
Based on the original CCTA examination reports, stenosis information was extracted for the RCA, LAD, LCX, and LM. Under the supervision of senior radiologists, the free-text imaging diagnostic reports were structured, cleaned, and standardized. The extracted coronary artery findings were then mapped to vessel-specific stenosis categories. According to clinically accepted standards, luminal stenosis was categorized as normal (<25\%), mild stenosis (25-50\%), moderate stenosis (50-70\%), or severe stenosis ($\geq 70\%$).

For binary label construction in model development and evaluation, the positive class was defined in a vessel-specific manner to identify hemodynamically significant stenosis. Specifically, a positive lesion was defined as luminal stenosis of $\geq 70\%$ for the RCA, LAD, and LCX. For the LM, the positive threshold was set at $\geq 50\%$. All luminal narrowing below these vessel-specific thresholds was grouped as the negative class. This tailored binary framework was applied independently to each coronary artery territory.

Simultaneously, free-text ECG diagnostic reports were standardized and recoded to classify the study population into normal and abnormal ECG groups, enabling stratified assessment of the model’s ability to detect occult coronary lesions in individuals without overt ECG abnormalities. Normal ECGs were identified using a rule-based natural language processing pipeline. Reports were classified as normal if they explicitly contained standard descriptors of normal findings, such as “normal ECG” or “no abnormality detected,” or if they met a stringent baseline criterion consisting of documented sinus rhythm and the absence of predefined pathological terms, including hypertrophy, conduction block, infarction, premature beats, ischemia, high voltage, or injury. To avoid misclassification caused by negated abnormal findings, such as “no obvious ST-T changes,” a regular-expression-based whitelist was applied before pathological keyword screening to identify and neutralize negative assertions. Reports that did not meet these predefined normal criteria were classified as abnormal ECGs. All automated classifications were subsequently reviewed by clinicians to ensure clinical validity and reliability.

All label generation, cleaning, and standardization procedures underwent manual auditing and confirmation by multiple senior clinicians to ensure consistency and reliability.

\subsection*{Model Architecture and Training Strategy}
This study fine-tuned the ECGFounder model trained for future new-onset MI using a transfer learning strategy \cite{ecgfounder1,ecgfounder2}. An AI-ECG model based on a multi-task deep learning framework was constructed to jointly predict hemodynamically significant stenosis of four major coronary arteries (RCA, LM, LAD, LCX), defined using vessel-specific thresholds. The model uses a Net1D network structure with 12-lead ECG signals as input. 

To address the gradient conflict and weight imbalance issues between different tasks during multi-task learning, an adaptive weighting strategy based on task uncertainty was introduced \cite{trainuncertain}. Specifically, for each task $t$, the loss function was weighted by a learnable uncertainty parameter $\sigma_t$, and the overall objective was defined as:

\begin{equation}
\mathcal{L} = \sum_{t=1}^{T} \left( \frac{1}{2\sigma_t^2} \mathcal{L}_t + \log \sigma_t \right),
\end{equation}

where $\mathcal{L}_t$ denotes the task-specific loss and $\sigma_t$ represents the task uncertainty, which is optimized jointly with model parameters. This formulation enables dynamic balancing of multiple tasks during training.

In addition, we used a projected-gradient conflict mitigation strategy inspired by PCGrad to reduce potential gradient interference among artery-specific tasks \cite{pcgrad}. For two task gradients $g_i$ and $g_j$, when their dot product is negative, indicating conflicting optimization directions, the gradient was adjusted as:

\begin{equation}
g_i \leftarrow g_i - \frac{g_i^\top g_j}{\|g_j\|^2} g_j, \quad \text{if } g_i^\top g_j < 0,
\end{equation}

thereby projecting conflicting gradients onto a non-conflicting direction to improve convergence stability.

The optimization process employed the AdamW optimizer combined with warmup and cosine decay strategies to dynamically adjust the learning rate, enhancing training robustness and generalization performance.

Data partitioning employed a five-fold stratified grouped cross-validation method, using patient ID as the grouping criterion to ensure that data from the same patient did not appear simultaneously in both the training and validation sets, while maintaining the balance of multi-label distribution. The model used the validation set Macro-AUC as the primary performance monitoring metric, and the weights of the best-performing model in each fold were stored.

After reading the 12-lead ECG signal, lead-by-lead z-score standardization was first performed to eliminate the impact of inter-individual amplitude differences on model training. To enhance the model’s robustness to signal perturbations encountered in real-world clinical scenarios, an online data augmentation strategy was employed during training to generate augmented ECG samples via controlled random perturbations.

Specifically, each ECG signal was subjected to stochastic transformations, including random temporal shifting, amplitude scaling, additive Gaussian noise injection, and local random occlusion. Temporal shifting was implemented by rolling the signal along the time axis within a range of ±10\% of the signal length, simulating beat misalignment. Amplitude scaling was performed by multiplying the signal with a random factor sampled from a uniform distribution in the range [0.9, 1.1], mimicking inter-beat voltage variation. Gaussian noise with zero mean and variance proportional to the augmentation intensity was added to emulate physiological and acquisition noise, while local random occlusion was applied by setting a contiguous segment (2\%–10\% of signal length) to zero, simulating signal dropout or artifact interference.

The augmentation strength was modulated by an epoch-dependent cosine annealing function. Specifically, we first defined a cosine decay factor as
\begin{equation}
s(e) = 0.5 \left( 1 + \cos\left( \frac{\pi e}{E} \right) \right),
\end{equation}
where $e$ denotes the current training epoch and $E$ represents the total number of training epochs. The final augmentation intensity was then computed as
\begin{equation}
\alpha(e) = 0.5 + 0.5 s(e),
\end{equation}
which is equivalent to
\begin{equation}
\alpha(e) = 0.75 + 0.25 \cos\left( \frac{\pi e}{E} \right).
\end{equation}
This formulation ensured that the effective augmentation intensity gradually decreased from 1.0 at the beginning of training to 0.5 at the final epoch. Specifically, the probability of applying time shifting, amplitude scaling, Gaussian noise injection, and random temporal masking was scaled by $\alpha(e)$. For Gaussian noise injection, the noise magnitude was also multiplied by $\alpha(e)$, whereas the predefined ranges of temporal shifting, amplitude scaling, and masking length remained unchanged. This strategy allowed the model to learn robust invariant representations under stronger perturbations during early training and then progressively focus on more stable ECG morphological features in later stages, thereby improving generalization while reducing the risk of excessive distortion of physiologically relevant ECG patterns.

During training, both the original ECG signal and its augmented counterpart were simultaneously fed into the network, encouraging the model to learn consistent representations across perturbed signal variations.

The training strategy described above, while ensuring data isolation at the patient level, enables the model to make stable joint predictions of multivascular lesions and further outputs continuous probability results, providing a methodological basis for subsequent risk stratification analysis, calibration assessment and clinical decision support.

\subsection*{Evaluation Methods}
To systematically evaluate model performance, we plotted receiver operating characteristic (ROC) curves and calculated the 95\% confidence interval (CI) for each area under the curve (AUC), thereby quantifying the classification and discrimination ability of the model. For internal validation, AUC was computed using a micro-averaging strategy, in which predictions from all five cross-validation folds were pooled before evaluation. While this internal micro-averaged AUC provides a baseline estimate of model learning capability, we recognize that utilizing these cross-validation folds for epoch selection inherently introduces a mild optimization bias. To rigorously address this and establish the true out-of-sample performance, an independent external validation dataset was utilized as a strictly held-out test set. For the external validation dataset and the subgroup of ECGs interpreted as normal in routine clinical practice, we adopted an ensemble strategy to obtain robust estimates. Specifically, each sample was independently processed by the five models saved during 5-fold cross-validation. The predicted probabilities from these models were then averaged to generate a final consensus probability for AUC calculation. And the five fold-models contributed with equal weights. The external validation dataset was strictly independent and was not used during training, hyperparameter tuning, or fold-wise model selection. In addition, we performed analyses at both the examination and patient levels and generated the corresponding ROC curves to assess whether model performance remained stable across different analytic units.

To examine whether model outputs reflected the anatomical severity of coronary artery disease, we visualized predicted probability distributions across CCTA-defined diagnostic categories using jittered strip plots, with overlaid medians and percentile-based error bars. We further used Spearman rank correlation analysis to assess the trend relationship between coronary stenosis severity and model-predicted probability. Specifically, the severity of stenosis is coded into ordered numerical categories, including no significant stenosis (stenosis $\leq25\%$), mild stenosis (stenosis 25\%-50\%), moderate stenosis (stenosis 50\%-70\%), severe stenosis or greater stenosis (stenosis $\geq70\%$). The monotonic association between these ordered categories and the model outputs was then evaluated to determine whether predicted probabilities increased with stenosis severity, thereby assessing model discrimination and reliability from the perspective of probability distribution.

We also performed subgroup analyses by stratifying the population according to age ($<65$ vs. $\geq65$ years), sex, prior medical history, the time interval between ECG and coronary CT angiography (CCTA) examination ($\leq3$ hours vs. $>3$ hours), medication records, clinical symptoms, presence of calcified plaque, and arrhythmia type. Model performance was systematically compared across these subgroups to evaluate its robustness under different demographic, clinical, and temporal conditions. 

\subsection*{Clinical Risk Stratification}
For each coronary artery prediction task, we converted the continuous model-predicted probability into three clinically interpretable risk categories: low, intermediate, and high risk. Two vessel-specific operating thresholds were selected for each model. The lower threshold defined the boundary between the low- and intermediate-risk groups, whereas the upper threshold defined the boundary between the intermediate- and high-risk groups.

The low-risk threshold was selected under a prespecified sensitivity constraint of 90\%. For a given candidate threshold \(t\), patients with predicted probability \(p \geq t\) were classified as test-positive, and those with \(p < t\) were classified as test-negative. The threshold-specific confusion matrix was defined as
\[
TP_t = \#(p \geq t, y=1), \quad FP_t = \#(p \geq t, y=0),
\]
\[
TN_t = \#(p < t, y=0), \quad FN_t = \#(p < t, y=1)
\]
where \(y=1\) denotes clinically significant vessel-specific stenosis, defined as \(\geq 70\%\) stenosis for RCA, LAD, and LCX and \(\geq 50\%\) stenosis for LM. Accordingly, \(y=0\) denotes non-significant stenosis under the corresponding vessel-specific definition. Sensitivity at threshold \(t\) was calculated as
\begin{equation}
Sensitivity_t = \frac{TP_t}{TP_t + FN_t}.
\end{equation}
The low-risk threshold \(t_{\mathrm{low}}\) was defined as the probability threshold satisfying the target sensitivity of 90\%. Patients with predicted probabilities below \(t_{\mathrm{low}}\) were assigned to the low-risk group. This threshold was designed to support rule-out decisions. The corresponding negative predictive value was calculated from the confusion matrix at \(t_{\mathrm{low}}\) with Jeffreys smoothing to improve numerical stability:
\begin{equation}
NPV_{\mathrm{low}} = \frac{TN_{t_{\mathrm{low}}}+0.5}{TN_{t_{\mathrm{low}}}+FN_{t_{\mathrm{low}}}+1.0}.
\end{equation}
This continuity correction was applied to reduce instability when event counts were small.

The high-risk threshold was selected under a prespecified specificity constraint of 95\%. For each candidate threshold \(t\), specificity was calculated as
\begin{equation}
Specificity_t = \frac{TN_t}{TN_t + FP_t}.
\end{equation}
The high-risk threshold \(t_{\mathrm{high}}\) was defined as the probability threshold satisfying the target specificity of 95\%. Patients with predicted probabilities equal to or above \(t_{\mathrm{high}}\) were assigned to the high-risk group. This threshold was designed to support rule-in decisions. The corresponding positive predictive value was calculated from the confusion matrix at \(t_{\mathrm{high}}\), again using Jeffreys smoothing:
\begin{equation}
PPV_{\mathrm{high}} = \frac{TP_{t_{\mathrm{high}}}+0.5}{TP_{t_{\mathrm{high}}}+FP_{t_{\mathrm{high}}}+1.0}.
\end{equation}
Accordingly, patients were stratified into three risk groups according to the model-predicted probability:
\[
\mathrm{Low\ risk}: p < t_{\mathrm{low}},
\]
\[
\mathrm{Intermediate\ risk}: t_{\mathrm{low}} \leq p < t_{\mathrm{high}},
\]
\[
\mathrm{High\ risk}: p \geq t_{\mathrm{high}}.
\]
Importantly, the negative predictive value for the low-risk boundary and the positive predictive value for the high-risk boundary were derived from two distinct threshold-specific confusion matrices, rather than from a single shared operating point. This design allows the low-risk threshold to prioritize safe exclusion of clinically relevant coronary stenosis, while allowing the high-risk threshold to prioritize reliable identification of patients with a high probability of disease.

All vessel-specific thresholds were derived from the internal validation set and then applied unchanged to the external validation set for evaluation, without any threshold re-optimization on the external data.

To evaluate the clinical validity of the model-derived risk strata, we generated calibration curves based on the stratified predictions and assessed the agreement between predicted risk and observed disease frequency. We further performed decision curve analysis to quantify the net clinical benefit of the model across a range of risk thresholds and compared it with treat-all and treat-none strategies.

For longitudinal prognostic assessment, vessel-level risk strata were further aggregated into patient-level risk categories in the follow-up cohort. For each coronary artery, model-predicted probabilities were first converted into low-, intermediate-, and high-risk categories using the same predefined vessel-specific thresholds. At the patient level, individuals were classified as high risk if any of the four coronary arteries was assigned to the high-risk group, classified as low risk only if all four arteries were assigned to the low-risk group, and classified as intermediate risk otherwise. Major adverse cardiovascular events were defined as a composite endpoint including myocardial infarction, all-cause death, percutaneous coronary intervention, or coronary artery bypass grafting. The time-to-event was defined as the earliest occurrence time among these events, whereas patients without observed events were censored at the last available follow-up time. Follow-up times beyond the prespecified 800-day window were administratively censored. Kaplan-Meier analysis was performed to estimate the cumulative incidence of major adverse cardiovascular events across the three patient-level risk groups, and group differences were assessed using the log-rank test. Pairwise hazard ratios were estimated using Cox proportional hazards models with the high-risk group as the reference group, and group differences were assessed using the log-rank test.

We then developed a clinical decision scheme by integrating the AI-derived risk strata with the PTP score recommended by the ESC guidelines \cite{19guideline}. According to the guideline-based PTP categories, patients were classified as low risk when PTP was \(\leq5\%\), intermediate risk when PTP was between 5\% and 15\%, and high risk when PTP was \(\geq15\%\). The fusion strategy combined the three-level PTP category with the three-level AI-derived risk category to generate a final clinical risk group. When PTP indicated low risk, patients were classified as final low risk if the AI model predicted low or intermediate risk, whereas patients with AI-predicted high risk were assigned to the gray zone for further clinical assessment. When PTP indicated intermediate risk, patients were classified as final low risk if the AI model predicted low risk, assigned to the gray zone if the AI model predicted intermediate risk, and classified as final high risk if the AI model predicted high risk. When PTP indicated high risk, patients were assigned to the gray zone if the AI model predicted low risk and classified as final high risk if the AI model predicted intermediate or high risk.

After applying this fusion strategy, we evaluated its clinical performance by calculating the negative predictive value for the final low-risk group, the positive predictive value for the final high-risk group, and the proportion of patients assigned to the gray zone. We further calculated the net reclassification improvement of the fusion strategy relative to guideline-based PTP stratification to quantify its incremental value over the clinical reference approach.

\subsection*{Results' Explanation}
To investigate the potential electrophysiological basis of model-derived risk assessment, patients were stratified according to the predefined dual-threshold strategy, and ECG waveform characteristics were compared across risk groups. Model attribution analyses were further performed to interpret the predictions of the AI model and to identify ECG regions contributing to high-risk classification.

For each coronary artery task, the predicted probability generated by the model was first obtained together with the corresponding low-risk and high-risk thresholds. For a given vessel, samples with predicted probabilities below the low-risk threshold were assigned to the low-risk group. Samples with predicted probabilities greater than or equal to the low-risk threshold and lower than the high-risk threshold were assigned to the intermediate-risk group. Samples with predicted probabilities greater than or equal to the high-risk threshold were assigned to the high-risk group.

Single-beat ECG segments were then automatically extracted and aligned from the raw 12-lead ECG recordings of each patient. R-peak localization was performed using a peak detection-based beat alignment strategy. Lead II was selected as the reference lead, and candidate R peaks were identified using a local maximum detection algorithm. To reduce interference from noise fluctuations, baseline drift, and non-QRS components, only local peaks with amplitudes greater than 30\% of the maximum amplitude in lead II were retained. A minimum distance of 0.4 s was also required between adjacent peaks. At a sampling frequency of 500 Hz, this interval corresponded to 200 sampling points and was used to avoid misclassifying multiple local fluctuations or closely spaced pseudo-peaks within the same cardiac cycle as independent R peaks. Local peaks satisfying both the amplitude threshold and the minimum distance constraint were defined as R-peak locations.

Based on the detected R-peak positions, each cardiac cycle was temporally aligned and cropped into a standardized beat segment. Specifically, for each detected R peak, the beat segment was extracted from 0.25 s before that R peak to 0.65 s after the same R peak, thereby covering the major waveform components before and after ventricular depolarization within one cardiac cycle. Beat segments were discarded if the R peak was too close to the signal boundary to allow complete extraction of this predefined window. Signal samples containing non-finite values or abnormal shapes were also excluded to ensure the reliability of subsequent waveform statistics.

After beat extraction, each of the 12 leads in each beat segment was independently normalized using z-score normalization. Specifically, for each lead within a single beat segment, the mean and standard deviation were calculated within the extracted time window, and the lead signal was transformed to a zero-mean, unit-variance waveform. This procedure was used to reduce the influence of inter-individual amplitude differences, lead contact variability, and recording scale differences, thereby allowing comparisons across risk groups to focus on relative waveform morphology rather than absolute signal magnitude.

For visualization, all beat waveforms from patients within each risk group were first aggregated, and the group-level mean beat waveform was then calculated for the low-, intermediate-, and high-risk groups. These mean waveforms were overlaid within the same coordinate system to visualize risk-associated morphological differences across the entire cardiac cycle. To further quantify global waveform differences between risk strata, lead-wise root mean square differences were calculated for the intermediate-risk group versus the low-risk group and for the high-risk group versus the low-risk group. Specifically, for each coronary artery task and each ECG lead, the RMS difference between the two group-level mean waveforms was computed across the entire beat window. This metric captured the overall magnitude of waveform morphology shift between risk groups and was displayed as a heatmap across coronary artery tasks and ECG leads.

Model attribution analysis using Integrated Gradients (IG) was performed to identify the key ECG temporal regions contributing to high-risk predictions. IG was selected because it provides robust, axiomatically justified feature-level attribution by accumulating the gradients of the model's prediction output with respect to the input features along a linear interpolation path from a baseline signal (a zero-voltage flatline) to the actual ECG input. For the group-level attribution map, the top 50 samples with the highest predicted probabilities were selected for each vessel-specific task. Following R-peak alignment, the IG attribution scores for these high-risk samples were computed and averaged to obtain the mean temporal distribution of model attention. 

\section*{Data Availability}
The data that support the findings of this study are not publicly available due to restrictions imposed by institutional ethics committees and data governance policies of the participating hospitals. Access to the data may be considered upon reasonable request to the corresponding author, subject to approval by the relevant ethics committees.

\section*{Code Availability}
The code used for model development and evaluation in this study is publicly available at \url{https://github.com/shunxio/AnyECG-CCTA}.

\section*{Declaration statements}

\subsection*{Acknowledgements}
This work was supported by the Beijing Natural Science Foundation (QY26080), Beifimg Nowa Program(202604841257), National Natural Science Foundation of China (62102008, 82470527), CCF-Tencent Rhino-Bird Open Research Fund (CCF-Tencent RAGR20250108), CCF-Zhipu Large Model Innovation Fund (CCF-Zhipu202414), PKU-OPPO Fund (BO202301, BO202503), Research Project of Peking University in the State Key Laboratory of Vascular Homeostasis and Remodeling (2025-SKLVHR-YCTS-02), Beijing Municipal Science and Technology Commission (Z251100000725008), Prevention and Control of Emerging and Major Infectious Diseases-National Science and Technology Major Project (2025ZD01906000, 2025ZD01906004), Capital’s Funds for Health Improvement and Research (CFH2026-1-4092).

The authors thank all collaborators and participating institutions for their support and contributions to this research.

\subsection*{Author Contributions}
Yujie Xiao led the conceptualization of the study, designed the methodology, implemented the models, performed validation and formal analyses, and drafted the initial version of the manuscript. Qinghao Zhao contributed to the manuscript revision process, assisted in refining the explanation of results, data cleaning and auditing, and provided critical feedback on the study. Hao Zhang and Tong Liu contributed to data acquisition, preprocessing, curation, cleaning and auditing. Gongzheng Tang, Zhuoran Kan, Deyun Zhang, Jun Li, Guangkun Nie, Xiaocheng Fang, Haoyu Wang, and Shun Huang contributed to data preparation and investigation. Jian Liu, Kangyin Chen provided resources, clinical expertise and contributed to data interpretation. Shenda Hong supervised the entire project, provided resources, guided study design and interpretation, served as the corresponding author, and takes full responsibility for the integrity of the work.  
All authors reviewed, revised, and approved the final manuscript.

\subsection*{Competing Interests}
Shenda Hong is an Associate Editor of \textit{npj Digital Medicine}. Shenda Hong was not involved in the journal’s review of, or decisions related to, this manuscript. The other authors declare no competing financial or non-financial interests.

\bibliography{reference}

\bigskip

\end{document}